\pdfoutput=1

\documentclass[11pt]{article}

\usepackage[]{acl}

\usepackage{times}
\usepackage{latexsym}
\usepackage{comment}
\usepackage{amsthm}

\usepackage[T1]{fontenc}

\usepackage[utf8]{inputenc}

\usepackage{microtype}

\usepackage{graphicx}

\title{WiC = TSV = WSD: \\ 
On the Equivalence of Three Semantic Tasks}

\author{
 Bradley Hauer,
 Grzegorz Kondrak\\
 Alberta Machine Intelligence Institute, Department of Computing Science\\
 University of Alberta, Edmonton, Canada\\
 {\tt {\{bmhauer,gkondrak\}}@ualberta.ca}
}

\date{}

\newcommand{\focusword}{focus word}
\newcommand{\contextsentence}{context}
\newcommand{\thesis}{hypothesis}

\newtheorem{myobs}{Proposition}
\newtheorem{mycorollary}{Corollary}

\begin{document}

\maketitle

\begin{abstract}

The Word-in-Context (WiC)
task has attracted considerable attention in the NLP 
community,
as demonstrated by the popularity of the recent MCL-WiC 
\mbox{Sem}\mbox{Eval} shared task.
Systems and lexical resources from
word sense disambiguation (WSD)
are often used for the WiC task and WiC dataset construction.
In this paper,
we establish 
the exact relationship between WiC and WSD, as well as  
the related task of target sense verification (TSV).
Building upon 
a novel hypothesis on the equivalence of
sense and meaning distinctions,
we demonstrate
through the application of tools from theoretical computer science
that these three semantic classification problems 
can be pairwise reduced to each other,
and therefore are equivalent.
The results of experiments
that involve systems and datasets for both WiC and WSD
provide strong empirical evidence 
that our problem reductions work in practice.

\end{abstract}

\section{Introduction}
\label{intro}

This paper answers an open question 
about the the relation between two important tasks in lexical semantics.
Word sense disambiguation (WSD) is the task of 
tagging a word in context with its sense \cite{navigli2009}.
The word-in-context (WiC) problem is the task of deciding whether 
a word has the same meaning in two different contexts \cite{pilehvar2019}.
A crucial difference between the two tasks is that WSD depends on 
a pre-defined sense inventory\footnote{For the purposes of this paper, 
we assume that the WSD sense inventory,
the discrete enumeration of the senses of each content word,
is the WordNet sense inventory \cite{fellbaum98book},
which is a standard practice in WSD \cite{raganato2017}.}
while WiC does not involve any identification or description of word meanings.
Despite ongoing interest in both tasks, 
there is substantial disagreement in the literature 
as to whether WiC is a re-formulation of WSD 
(e.g. \newcite{levine2020})
or an entirely distinct task 
(e.g. \newcite{martelli2021}).

By establishing that WSD and WiC are equivalent,
we construct a theoretical foundation 
for the transfer of resources and methods between the two tasks.
WSD has been intensively studied for decades, while WiC 
has recently attracted considerable attention from the research community. 
For example, 
the MCL-WiC SemEval shared task \cite{martelli2021}
attracted 48 teams,
and WiC instances have been integrated into the SuperGLUE benchmark 
\cite{wang2019}.
Understanding how the two tasks relate to each other
allows us to correctly interpret and confidently build upon those results,
including prior work on using WSD systems for WiC
(e.g. \newcite{loureiro2019liaad}).

We establish the theoretical equivalence of WiC and WSD
by specifying reduction algorithms
which produce a solution for one problem 
by applying an algorithm for another.
In particular, 
we employ the target sense verification (TSV) task \cite{breit2021}
as an intermediate step between WSD and WiC, 
and specify three reductions: WiC to WSD, WSD to TSV, and TSV to WiC.
We formalize the three problems using a common notation, 
and provide both theoretical and empirical evidence 
for the correctness of our reductions.
While we focus on English in this paper, 
we make no language-specific assumptions.\footnote{\newcite{hauer2021semeval} 
leverage translations from multiple languages for the WiC task
by applying the substitution test for the
synonymy of senses \cite{hauer2020set}.}

The soundness of all three tasks 
hinges on the consistency of judgments of sameness of word meaning, 
whether with respect to discrete sense inventories (WSD), 
a representation of a single sense (TSV), 
or two occurrences of a word (WiC).
We posit that
{\em different instances of a word have the same meaning 
if and only if they have the same sense.}
This empirically falsifiable proposition, which we refer to as 
the {\em sense-meaning hypothesis}, implies that
WiC judgements induce sense inventories 
that correspond to word senses.
This counter-intuitive finding has
intriguing implications for the task of word sense induction (WSI), 
as well as algorithmic wordnet construction.

We empirically validate our hypothesis 
by conducting multiple experiments and analyzing the results.
In particular,
we test our WSD-to-WiC and WiC-to-WSD reductions
on standard benchmark datasets using state-of-the-art systems.
We find that our reductions perform remarkably well,
revealing no clear counter-examples to our hypothesis in the process.

Our contributions are as follows:
(1)
We answer the open question of the relation between WiC and WSD
by constructing a theoretical argument 
for their equivalence,
which is based on a novel sense-meaning hypothesis.
(2) We carry out a series of validation experiments 
that strongly support the correctness of our reductions.
(3) We release the details of our manual analysis and annotations 
of the instances identified in the validation experiments.

\section{Theoretical Formalization}
\label{sec:theory}

In this section, 
we formally define the three problems,
present a theoretical argument for their equivalence,
and specify the reductions.

\subsection{Problem Definitions}
\label{sec:defs}

Senses in our problem definitions refer to {\em wordnet senses}.
A {\em wordnet} is a theoretical construct
which is composed of synonym sets, or \emph{synsets},
such that each synset corresponds to a unique concept,
and each sense of a given word corresponds to a different synset.
Actual wordnets, such as Princeton WordNet \cite{fellbaum98book},
are considered to be imperfect implementations of the theoretical construct.

In the problem definitions below,
$C, C_1, C_2$ represent contexts,
each of which contains a single {\focusword} $w$ used in the sense $s$.
We assume that every content word token is used in exactly one 
sense.\footnote{This is empirically supported by the fact that
99.7\% of annotated tokens in SemCor
are assigned a single sense.}

\begin{itemize}
\item
{\bf \mbox{WSD}$(C, w)$}:
{\rm Given a {\contextsentence} $C$ which contains a single {\focusword} $w$, 
return the sense $s$ of $w$ in $C$.}

\item
{\bf \mbox{TSV}$(C, w, s)$}:
{\rm Given a {\contextsentence} $C$ which contains a single {\focusword} $w$, 
and a sense $s$,
return {\sc True}
if $s$ is the sense of $w$ in $C$,
and {\sc False} otherwise.}

\item
{\bf \mbox{WiC}$(C_1, C_2, w)$}:
{\rm Given two {\contextsentence}s $C_1$ and $C_2$ 
which contain the same {\focusword} $w$, 
return {\sc True}
if $w$ has the same meaning in both $C_1$ and $C_2$,
and {\sc False} otherwise.}

\end{itemize}

\subsection{Problem Equivalence}

The theoretical argument for 
the sense-meaning hypothesis
is based on the assumption that
the relation of sameness of word meaning
is shared between the three problems.
This is supported by the lack of 
distinction between {\em meanings} and {\em senses} in 
the original WiC task proposal.\footnote{{\rm
``The proposed dataset, WiC, is based on lexicographic
examples, which constitute a reliable basis to [\ldots]
discern different {\bf meanings of words}.''}
\cite{pilehvar2019}.}
On the other hand,
WordNet exhibits
a strict one-to-one correspondence
between distinct meanings, synsets, and concepts,
with each word sense corresponding to a specific synset.
This implies that senses are
ultimately grounded in sameness of meaning as well.\footnote{{\rm ``[Each]
synonym set represents one underlying lexical concept.  [\ldots] 
{\bf Word meaning} [refers] to the lexicalized concept 
      that a [word] form can be used to express.''}
\cite{miller1995acm}.}
Therefore, every word meaning distinction should correspond to 
a pairwise sense distinction.
Contrariwise, if two tokens of the same word 
express different concepts, their meaning must be different.
This equivalence also includes the TSV problem,
provided that 
the given sense of the {\focusword} corresponds to a single synset.

\subsection{Problem Reductions}
\label{reductions}

We now present the three problem reductions.
For our purposes, 
a \textbf{P-to-Q} reduction is 
an algorithm that,
given an algorithm for a problem \textbf{Q},
solves an instance of a problem \textbf{P} 
by combining the solutions of one or more instances of \textbf{Q}.

\begin{myobs}
WiC is reducible to WSD.
\end{myobs}

To reduce {\bf WiC to WSD},
we directly apply the sense-meaning hypothesis from Section~\ref{intro}
by assuming that the focus word has the same meaning in two contexts 
if and only if it can be independently tagged with the same sense 
in both contexts. 
Formally:
\begin{center}
\begin{math}
\mbox{WiC}(C_1, C_2, w) \Leftrightarrow
\mbox{WSD}(C_1, w) = \mbox{WSD}(C_2, w)
\end{math}
\end{center}

\begin{figure}[t]
   \includegraphics[width=\linewidth]{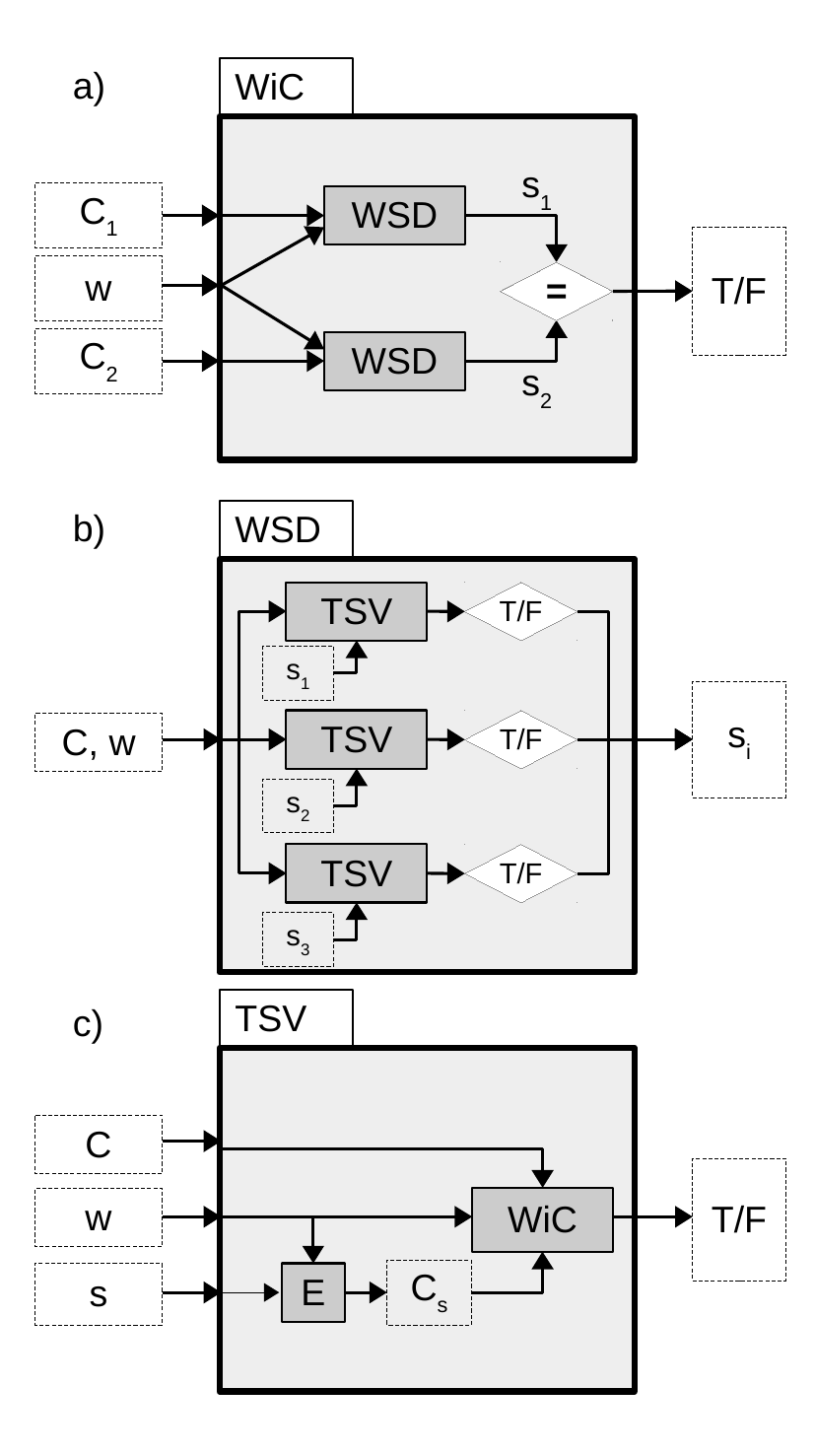}
   \caption{Three problem reductions:
a) WiC to WSD,
b) WSD to TSV,
and
c) TSV to WiC.}
   \label{fig:reductions}
\end{figure}

Thus, given a method for solving WSD, we can solve any given WiC
instance by solving the two WSD instances which consist of 
the focus word in the first and second context, respectively.
We return TRUE if the returned senses are equal, FALSE otherwise
(Figure~\ref{fig:reductions}a).

\begin{myobs}
WSD is reducible to TSV.
\end{myobs}

To reduce {\bf WSD to TSV},
we take advantage of the fact that TSV can be
applied to a variety of different sense representations, 
without any explicit dependence on a specific sense inventory. 
We can therefore query a TSV system with various senses of the focus word, 
using the same sense inventory as the WSD task:
\begin{center}
\begin{math}
\mbox{WSD}(C, w) = s 
\Leftrightarrow
\mbox{TSV}(C, w, s)
\end{math}
\end{center}

Thus, given a TSV solver, 
for any WSD instance we can construct a list of $k$ TSV instances,
one for each sense of the focus word in the corresponding WSD sense inventory.
We return the sense for which the TSV instance returns TRUE
(Figure~\ref{fig:reductions}b).
The correctness of this reduction hinges on the assumption
that every content word in context is used in exactly one sense.

\begin{myobs}
TSV is reducible to WiC.
\end{myobs}

To reduce {\bf TSV to WiC},
we again leverage our sense-meaning hypothesis
by assuming that 
a content word used in a particular sense
will be judged to have the same meaning as 
in an example sentence for that sense. 
Formally:
\begin{center}
\begin{math}
\mbox{TSV}(C, w, s)
\Leftrightarrow
\mbox{WiC}(C, C_s, w) 
\end{math}
\end{center}
where $C_s$ is a context in which $w$ is unambiguously used in sense $s$.
So, given a method for solving WiC, 
we can solve a TSV instance by
replacing the given sense representation with an example, yielding a
WiC instance
(Figure~\ref{fig:reductions}c).
This reduction depends on the existence
of an algorithm $E$ that,
given a sense $s$ of a word $w$,
can generate an example sentence $C_s$ 
that contains $w$ used in sense $s$.\footnote{This is related
to a well-defined and actively researched task
known as exemplification modelling
\cite{barba2021exemplification}.}

These three reductions are sufficient to establish the equivalence
of {\bf WSD}, {\bf TSV}, and {\bf WiC}.
A method which solves any of these problems
can be used to construct methods which solve the other two,
using a sequence of at most two of the above reductions.

In particular, we can reduce {\bf WSD to WiC}:
\begin{mycorollary}
WSD is reducible to WiC.
\end{mycorollary}
\label{wsd-to-wic-corollary}

To reduce {\bf WSD to WiC},
first reduce the WSD instance to TSV,
producing one TSV instance for each sense $s$ of $w$.
Then, reduce each of these TSV instances to a WiC instance,
by pairing the context of the WSD instance
with an example context for each sense.
Succinctly:
\begin{center}
\begin{math}
\mbox{WSD}(C, w) = s
\Leftrightarrow
\mbox{WiC}(C, C_s, w)
\end{math}
\end{center}

Thus, solving the original WSD instance
can be achieved by identifying 
the single positive instance
in the list of $k$ WiC instances.

\section{WiC Datasets} 
\label{sec:datasets}

In this section,
we discuss and analyze the existing WiC datasets
with the aim of finding a dataset suitable for 
validating our equivalence {\thesis}.
An instance that contradicts one of the reduction equivalences 
in Section~\ref{reductions}
would be an exception to the {\thesis}.
Since natural language is not pure logic,
falsifying the {\thesis} would require finding 
that such exceptions constitute a substantial fraction of instances,
excluding errors and omissions in lexical resources.

\subsection{WiC}
\label{sec:owic}

WiC was originally proposed as a dataset for the evaluation 
of contextualized embeddings, including neural language models 
\cite{pilehvar2019}.
The original WiC dataset
consists of
pairs of sentences drawn
mostly from WordNet,
which were further filtered to remove fine-grained sense distinctions.
The reported inter-annotator agreement was 80\% for the final pruned set,
and only 57\% for the pruned-out instances.

Since, regardless of the source,
all instances were annotated automatically 
by checking the sense identity in WordNet,
the WiC dataset cannot,
{\em by construction},
contain any exceptions to the equivalence {\thesis}.
Therefore, we do not use the original WiC dataset in our experiments.
Nevertheless, 
it is possible to automatically identify both senses 
in about half the instances in the dataset
by matching them to the sense usage example sentences in WordNet 3.0.
It is interesting to note
that combining such a WordNet lookup
with a random back-off on the remaining instances
results in 
correctly solving 76.1\% of the WiC instances in the test set,
which 
exceeds the current 
state-of-the-art of $72.1$\% \cite{levine2020}.

\subsection{WiC-TSV}
\label{sec:tsv}

\newcite{breit2021} propose 
\emph{target sense verification} (TSV), the task
of deciding whether a given word in a given context is used in a given sense.
TSV is similar to WiC in that it is also a binary classification task,
but only one context is provided.
TSV is also similar to WSD in that there is an explicit representation of senses,
but there is only one sense to consider.
Three sub-tasks are defined depending on the method of representing a sense:
(a) definition, (b) hypernyms, and (c) both definition and hypernyms.

Approximately 85\% of the instances in 
the WiC-TSV dataset
are derived directly from the original WiC dataset,
and so are ultimately based on WordNet senses.\footnote{Three smaller 
sets are devoted to cocktail, medical, and computer terms,
respectively,
and appear more related to named entity recognition than to WSD.}
Specifically, 
the sense of the focus word was established by reversing the process
by which the WiC instances were created,
as in the WordNet lookup procedure applied
to the WiC dataset in Section~\ref{sec:owic}.
Because of this construction method,
no exceptions to the equivalence {\thesis}
can be found in the WiC-TSV dataset. 

\subsection{MCL-WiC}
\label{sec:exp:mcl}

\newcite{martelli2021} introduce 
the Multilingual and Cross-lingual Word-in-Context dataset.
The English portion of the dataset 
consists of 10k WiC instances,
divided into a training set (8k instances),
as well as development and test sets (1k instances each).
The task is exactly the same as the original WiC task,
and matches our WiC problem formalization in Section~\ref{sec:defs}.
In particular,
while the dataset covers multiple languages,
the task itself remains monolingual,
in the sense that the system need only consider one language at a time;
that is, all input and output for a given instance is in a single language.

In contrast with the original WiC dataset,
which was largely derived from WordNet, 
the sentence pairs in MCL-WiC
were manually selected
and annotated.
Annotators
consulted ``multiple reputable dictionaries''
to minimize the subjectivity of their decisions
on the identity of meaning. 
As a result, both
the inter-annotator agreement
($\kappa = 0.968$),
and the best system accuracy (93.3\% on English, \newcite{gupta2021})
are much higher than 
on the original WiC dataset.

The MCL-WiC dataset (Section \ref{sec:exp:mcl}) is especially valuable
for testing our sense-meaning equivalence {\thesis}
because it does not rely on pre-existing WordNet sense annotations,
and is agnostic toward WordNet sense distinctions.
For this reason, we make the MCL-WiC dataset the focus 
of our empirical validation
experiments in the next section.

\section{Empirical Validation}
\label{sec:experiments}

In this section, 
we aim to quantify and analyze
any apparent counter-examples to the sense-meaning hypothesis
which are identified in the process of
testing the WSD-to-WiC and WiC-to-WSD reductions.
We are particularly interested in the exceptions
that cannot be attributed to errors in the resources 
that are used to implement the reductions,
because
such exceptions represent potential evidence against our hypothesis.

\subsection{Systems}
\label{wic-wsd-setup}

In order to implement the WSD-to-WiC and WiC-to-WSD reductions,
we adopt two recent systems designed for the WiC and WSD tasks, respectively.

Our WiC system of choice is LIORI \cite{davletov2021}.
In the MCL-WiC shared task, LIORI
obtained an accuracy of 91.1\% on the English {\em test set}, 
which was within 2\% of the best performing system.
LIORI works by concatenating each sentence pair into a single string,
and fine-tuning a neural language model for binary classification.
We use the code
made available by the 
authors\footnote{\url{https://github.com/davletov-aa/mcl-wic}},
and derive our model from the MCL-WiC English training set.
 
As our WSD system, we adopt ESCHER \cite{barba2021}.
ESCHER re-formulates WSD as a span extraction task:
For a given WSD instance, the context is concatenated with 
all glosses of the focus word into a single string,
from which the gloss of the correct sense is extracted.
We derive our model 
using the implementation and training procedure provided by the 
authors\footnote{\url{https://github.com/SapienzaNLP/esc}}.
The training data includes SemCor \cite{miller1993}.
In our replication experiments, this model achieves 80.1\% F1 
on the standard WSD benchmark datasets of \newcite{raganato2017}.

\subsection{Solving WSD with WiC}
\label{exp-wsd-wic}

Our first experiment involves
an implementation of the reduction of WSD to WiC.
For each WSD instance, we construct a set of WiC instances 
that correspond to its possible senses,
solve them with LIORI, and return a single sense,
in accordance with the reduction specified in
Corollary~1
from Section \ref{reductions}.
We then present and analyze the results on a standard WSD dataset.

\subsubsection{Implementation of the Reduction}

Given a WSD instance consisting of a focus word $w$ in a context $C$,
we create a set of $k$ WiC instances,
where $k$ is the number of senses of $w$.
In WordNet 3.0,
each sense $s$ has a gloss $g_s$,
and sometimes also a usage example of $w$ being used in sense $s$.
Since not all synsets are accompanied by usage examples,
we instead generate a new synthetic usage example $C_s$
for each sense of $w$
using the following pattern: 
$C_s :=$ ``\,`$w$' {\em in this context means} $g_s$''.
Thus $C_s$ represents an unambiguous example of $w$ being used in sense $s$.
The resulting WiC instance for $s$
is then composed of 
contexts $C$ and $C_s$, both of which include the focus word $w$.

Our LIORI model 
returns a binary classification
and a score
for each of the constructed WiC instances.
While LIORI may classify zero, one, or more instances as true,
our implementation returns only 
the sense with the highest score.
This is in accordance with the definition of the WSD task as
identifying a single correct sense for a word in context
(Section~\ref{sec:defs}).

\subsubsection{Results and Discussion}

To estimate the expected accuracy of the above implementation,
we first apply LIORI to 
the 1000 instances in the MCL-WiC English development set.
LIORI achieves an accuracy of 88.0\%,
which we use as an estimate of the probability
that LIORI correctly classifies any given WiC instance.
The average number of senses per instance 
in our WSD dataset
is approximately 8.5.
Since any error by LIORI can cause the WSD-to-WiC reduction 
to output the wrong sense,
we estimate the expected probability that LIORI correctly classifies 
a single WSD instance 
as ${0.880^{8.5} \approx 0.34}$.

We test the reduction on
the SemEval 2007 dataset, as provided by \newcite{raganato2017}.
This test set contains 455 WSD instances, 
all but four of which (over 99\%)
are annotated with exactly one sense.
Our reduction implementation obtains an accuracy of 47.9\%
by returning a single predicted sense for every WSD instance in the test set.
As this result is substantially higher than the expected accuracy of 34\%,
we interpret it as evidence in favor of our hypothesis.

In theory, 
for each WSD instance,
LIORI should classify as true exactly one of the constructed WiC instances,
which represents the single correct sense.
In practice, 
this is the case in only 48 out of 455 cases.
Our reduction implementation predicts the correct sense for 38 out of 48,
yielding a precision of 79.2\%.
We verified that ESCHER,
trained on over 226k sense annotations in SemCor,
correctly annotates 39 of these 48 instances. 
On this subset of instances, our WSD-to-WiC reduction based on LIORI
is therefore competitive with state-of-the-art supervised WSD systems,
despite not depending on any sense-annotated training data.
This constitutes further evidence for the correctness of our reduction,
and our hypothesis.

\subsection{Solving WiC with WSD}
\label{sec:exp:auto}

In this experiment, 
we apply a state-of-the-art supervised WSD system
to solve, via our WiC-to-WSD reduction,
all WiC instances in an independently-annotated test set.  
We then manually analyze a sample of the errors
to assess whether the experiment supports our hypothesis
and the correctness of our reduction.

\subsubsection{Implementation of the Reduction}

The implementation of the WiC-to-WSD reduction 
is conceptually simpler that the previously described WSD-to-WiC
reduction.\footnote{In fact, \newcite{loureiro2019liaad}
implicitly apply this reduction on a WiC dataset with their  
WSD system LMMS.}
Given a WiC instance consisting of contexts $C_1$ and $C_2$
for a word $w$,
we create two corresponding WSD instances: $(C_1,w)$ and $(C_2,w)$.
Both WSD instances are passed to ESCHER,
which independently assigns senses $s_1$ and $s_2$ to $w$ 
in each of the two contexts.
We classify the WiC instance as positive if and only if $s_1 = s_2$.

There are two types of possible counter-exam\-ples to our hypothesis:
(1) 
a WiC instance which is annotated as positive (i.e., the same meaning)
in which both focus tokens have different senses;
and (2)
a WiC instance which is annotated as negative (i.e., different meanings)
in which both focus tokens have the same sense.
These two types could arise from WSD sense distinctions that are
too fine-grained or too coarse-grained, respectively.

\subsubsection{Expected Accuracy}
\label{exp_acc}

The expected accuracy of the WiC-to-WSD reduction 
is more complex to calculate than that of the WSD-to-WiC reduction.
Our calculation is based on the simplifying assumption that 
all WSD errors are independent and equally likely.
For the probability that ESCHER disambiguates any WSD instance correctly,
we use the value of $p=0.801$,
based on our replication result in Section~\ref{wic-wsd-setup}.
The average number of senses per focus token 
in the dataset used in our experiment is 
$k=4.73$.
Since there are $k-1$ incorrect senses for each WSD instance,
we approximate 
the probability of predicting a given incorrect sense
in either WiC sentence as 
$q = (1-p)/(k-1) = 0.053$.

In order to estimate the probability of a correct classification,
we consider two main cases.
\begin{enumerate}
\item 
A {\em positive} WiC instance is {\em correctly} classified as positive if 
either (1.1) both corresponding WSD instances are disambiguated correctly,
or (1.2) both instances are tagged with the same incorrect sense:
$P_1 = p^2 + (k-1)q^2 = 0.642 + 0.011$.
\item 
A {\em negative} WiC instance is {\em incorrectly} classified as positive if
either (2.1) one of the corresponding WSD instances is disambiguated correctly 
and the other is incorrectly tagged with the same sense,
or (2.2) both instances are tagged with the same incorrect sense:
$P_2 = 2pq + (k-2)q^2 = 0.085 + 0.008$.
\end{enumerate}
Assuming that the dataset is balanced,
the expected
probability of classifying a WiC instance correctly is therefore:
$P_1/2 + (1 - P_2)/2 =$ \textbf{0.779}.

\subsubsection{Results and Discussion}
\label{wic-wsd-results}

We test the reduction on the MCL-WiC English development set,
which consists of 500 positive and 500 negative WiC instances.
We tokenize, lemmatize, and POS-tag all 2000 sentences
with TreeTagger\footnote{\url{https://cis.uni-muenchen.de/~schmid/tools/TreeTagger}}
\cite{schmid1999} as a pre-processing step.
ESCHER is then applied to predict the sense of the focus word in each 
sentence.
In 25 cases, 
ESCHER failed to make a sense prediction,
that is, one or both focus words were not disambiguated,
due to TreeTagger tokenization or lemmatization errors. 
The accuracy on the remaining 975 instances is 
78.5\%,
which is within 1\% of our theoretical estimate in Section~\ref{exp_acc}.
We conclude that
this experiment provides
strong empirical support for our hypothesis
and the correctness of our reductions.

\subsubsection{Analysis}
\label{wic-wsd-analysis}

To further evaluate our WiC-to-WSD reduction,
we manually analyzed a sample of 10 false positives and 10 false negatives
from this experiment.
The sample was {\em not} random;
instead, we attempted to automatically 
select the instances that were most likely to
represent exceptions to our equivalence {\thesis}.
Specifically, we restricted the analysis
to WiC instances that were {\rm correctly} classified by LIORI,
in order to reduce the impact of 
erroneous annotations, which are unavoidable in any gold dataset.
As a result,
the accuracy of ESCHER on the WSD instances in this sample
is expected to be lower than in the entire dataset.
In fact,
in 13 of the 20 instances (six false positives, seven false negatives), 
the misclassification was due to an error made by ESCHER.

In three of the seven remaining cases (all false positives),
the WiC misclassification was caused by
the WordNet sense inventory not including the correct sense
of one of the focus tokens.
Since we 
require ESCHER to produce a WordNet sense as output,
such omissions preclude the correct disambiguation of the focus word.
In all such cases, we were able to find the omitted sense in one of
the dictionaries that we consulted (Oxford or Merriam-Webster).
For example,
the correct sense of the verb {\em partake} in the WiC sentence
``he has \textbf{partaken} in many management
courses''
is ``join in (an activity)''
which is in the Oxford English Dictionary, but not in WordNet 3.0.
The missing WordNet senses 
for each of these instances
are shown in rows 1-3 of Table~\ref{tab:missing-wn-senses}.
 
Among the remaining four instances,
in one anomalous case 
we disagreed on the WordNet sense
of the adverb {\em richly} in the phrase 
{\em richly rewarding}.
However, in the other three cases,
ESCHER's annotations were unquestionably correct. 
We defer the discussion of those three interesting instances
to the next section. 

\subsection{Manual Annotation Experiment}
\label{sec:exp:annot}

To further expand our analysis,
we manually analyzed 60 additional randomly selected instances
from the English MCL-WiC training set.
The size of the sample was limited
because WSD instances are difficult and time-consuming to analyze,
especially when multiple annotators are involved
and an effort is made to avoid any unconscious bias.

For each such instance, 
we assigned WordNet senses to each of the two focus tokens,
without accessing the gold MCL-WiC labels.
Our judgments were
based on the glosses and usage examples of the available senses,
as well as the contents of the corresponding synsets
and their hypernym synsets.
Subsequently, we analyzed each instance where the WiC prediction 
obtained by applying the WiC-to-WSD reduction 
did not match the WiC classification 
in the official gold data.\footnote{We publish the annotated 
set of 60 WiC instances at
\url{https://webdocs.cs.ualberta.ca/\~kondrak}}

We found that 55 out of 60 instances (91.7\%) unquestionably
conform to the equivalence {\thesis}.
The remaining five instances
can be divided into three categories:
(1) tokenization errors in MCL-WiC,
(2) missing senses in WordNet,  and
(3) possible annotation errors in MCL-WiC.
We discuss these three types of errors below.

\begin{table}[t]
\begin{center}
\begin{small}
\begin{tabular}{|c|l|p{33mm}|c|}
\hline
 & Lemma & Gloss & Dict \\
\hline
1 & partake (v) & join in (an activity) & OED \\
\hline
2 & instant (adj) & prepared quickly and with little effort & OED \\
\hline
3 & familiar (adj) & of or relating to a family & MW  \\
\hline
4 & breach (v) & to leap out of water & MW \\ 
\hline
5 & spotter (n) & a member of a motor racing team &  OED \\
\hline
\hline
6 & campaign (n) & an organized course of action to achieve a goal & OED \\
\hline
7 & campaign (n) & a set of organized actions that a political candidate undertakes in 
an election & OED \\
\hline
8  & drive (n) & determination and ambition to achieve something & OED \\
\hline
9  & drive (n) & an organized effort by a number of people & OED \\
\hline
10  & wedding (n) & a marriage ceremony with accompanying festivities & MW \\
\hline
11  & wedding (n) & an act, process, or instance of joining in close association & MW \\
\hline
12 & analyst (n) & someone who analyzes & Wik \\
\hline
13 & analyst (n) & a financial analyst; a business analyst & Wik \\
\hline
\end{tabular}
\caption{Examples of senses that are not in WordNet
(Rows 1-5),
and sense distinctions found in external dictionaries
(Rows 6-13):
OED (Oxford English Dictionary), MW (Merriam-Webster), Wik (Wiktionary).
}
\label{tab:missing-wn-senses}
\end{small}
\end{center}
\end{table}

In two instances, word tokenization errors 
interfere with the MCL-WiC annotations:
(1) {\em together} in 
``the final {\bf coming together}''
is annotated as an adverb instead of a particle 
of a phrasal verb,
and 
(2) {\em shiner} in 
``{\bf shoes shiners} met the inspector''
is annotated as a stand-alone noun instead of a part of
a compound noun.
These tokenization errors prevent the proper assignment of WordNet senses.

In two instances 
(rows 4 and 5 in Table~\ref{tab:missing-wn-senses}), 
one of the senses of the {\focusword} is missing in WordNet:
(1) {\em breach} referring to 
an animal breaking through the surface of the water, 
and 
(2) {\em spotter} referring to
a member of a motor racing team 
who communicates by radio with the driver.
Neither of these senses is subsumed by another sense in WordNet,
and both of them are present in one of the consulted dictionaries.

In the final problematic instance,
MCL-WiC classifies
the noun {\em campaign} as having the same meaning in the contexts
``{\rm during the election {\bf campaign}}'' and
``{\rm the {\bf campaign} had a positive impact on behavior.}''
Since
the distinction between these two senses of {\em campaign} is found in
the Oxford English Dictionary,
which was among the ones consulted by the MCL-WiC annotators \cite{martelli2021},
we classify it as an MCL-WiC annotation error
(rows 6 and 7 in Table~\ref{tab:missing-wn-senses}). 

Similarly,  
we posit an MCL-WiC annotation error in each of
the three outstanding false negatives from Section \ref{wic-wsd-analysis},
which could not be attributed to ESCHER,
based on the verification in external dictionaries.
For example,
unlike WordNet,
Oxford and Merriam-Webster both distinguish
the emotional and organizational meanings of {\em drive}.
Similar analysis applies in instances involving the words
{\em wedding}
and
{\em analyst}
(rows 8-13 in Table~\ref{tab:missing-wn-senses}).
Since the meanings of the focus words in these contexts
are distinguished in a dictionary,
they should be considered distinct meanings
according to the annotation procedure of 
\newcite{martelli2021}.
We conclude that in these cases,
the MCL-WiC label is incorrect, 
and so they do not constitute exceptions to our hypothesis.

In summary,
a careful analysis 
of 25 apparent exceptions made by our reduction across 80 instances,
using both automatic and manual WSD,
reveals no clear 
evidence against the correctness of our reduction.
We therefore conclude that
the results of these experiments
strongly support our {\thesis}.

\section{Discussion}
\label{sec:discuss}

Having presented theoretical and empirical evidence for the equivalence
of WiC, WSD, and TSV,
we devote this section to the discussion of the relationship between
WordNet and WiC.

Most English WiC and TSV datasets
are based, in whole or in part, on WordNet.
If no sense inventory is used
for grounding decisions about meaning,
the inter-annotator agreement is reported to be only about 80\%
\cite{pilehvar2019,breit2021}.
For the MCL-WiC dataset, however, annotators consulted other dictionaries,
and obtained ``almost perfect agreement" \cite{martelli2021}. 
This suggests that sense inventories, and semantic resources in general,
are crucial to reliable annotation for semantic tasks.
However, because 
the exact MCL-WiC procedure for resolving differences between dictionaries 
is not fully specified,
and because such dictionaries vary in their availability,
the correctness of the annotations cannot be readily verified
(c.f. Section~\ref{sec:exp:annot}).

Our experiments provide evidence that, 
even when the WordNet sense inventory 
is not explicitly used in constructing WiC datasets,
WiC annotations nevertheless tend to agree with WordNet sense distinctions,
as our {\thesis} predicts.
Namely, the MCL-WiC instances in which both focus tokens have the same sense 
are almost always annotated as positive
by the MCL-WiC annotators.
The converse also holds, 
with any exceptions being explainable by errors in the resources.
Thus, empirical validation
confirms our sense-meaning \thesis{},
which implies that the meaning distinctions induced by WiC judgements 
closely match WordNet sense inventories.
This is a remarkable finding
given the high granularity of WordNet.

We postulate that
the adoption of WordNet as the standard sense inventory for WiC
would have several practical benefits:
(1) it has been adopted as the standard inventory for WSD,
and so would simplify multi-task evaluation;
(2) it allows seamless application of systems across datasets;
(3) it facilitates rapid creation of new WiC datasets
based on existing sense-annotated corpora;
(4) it is freely available;
(5) it can be modified and extended to correct errors and omissions
\cite{mccrae2020};
and finally
(6) it can be extended to facilitate work with other languages,
as in the XL-WiC dataset \cite{raganato2020}.

In addition, WordNet has strong theoretical advantages.
Its fine granularity is a consequence of
its grounding in synonymy and lexical concepts.
Therefore, the sense distinctions found in other dictionaries 
either already correspond to different WordNet concepts,
or should lead to adding new concepts to WordNet.
Furthermore,
unlike in dictionaries,
senses of different words in WordNet
are linked via semantic relations such as synonymy and hypernymy,
which facilitate an objective assignment
of every word usage to a single WordNet concept.
This property of WordNet may be the reason that
the WSD methods 
based on sense relation information
have surpassed the inter-annotator agreement ceiling of around 70\%
\cite{navigli2006}.

\section{Conclusion}
\label{sec:conclusion}

We formulated a novel sense-meaning {\thesis},
which allowed us to demonstrate 
the equivalence of three semantic tasks
by mutual reductions.
We corroborated our conclusions 
by performing a series of experiments 
involving both WSD and WiC tools and resources.
We have argued that
these relationships originate from the WordNet properties,
which are highly desirable in semantics research.
We expect that our findings will stimulate future work on
system development, resource creation, and
joint model optimization
for these tasks.

\section*{Acknowledgements}

This research was supported by
the Natural Sciences and Engineering Research Council of Canada (NSERC),
and the Alberta Machine Intelligence Institute (Amii).

\bibliography{wsd.bib}

\end{document}